\documentclass[letterpaper]{article} 
\usepackage{aaai23}  
\usepackage{times}  
\usepackage{helvet}  
\usepackage{courier}  
\usepackage[hyphens]{url}  
\usepackage{graphicx} 
\usepackage{natbib}  
\usepackage{caption} 
\usepackage{algorithm}
\usepackage{algpseudocode}
\usepackage{algorithmicx}

\usepackage{booktabs}
\usepackage{amsfonts} 
\usepackage{amsmath}
\usepackage{xcolor}

\usepackage{mathtools}
\usepackage{amssymb}
\usepackage{amsthm,bm} 

\usepackage{newfloat}
\usepackage{listings}
\DeclareCaptionStyle{ruled}{labelfont=normalfont,labelsep=colon,strut=off} 
\floatstyle{ruled}
\newfloat{listing}{tb}{lst}{}
\floatname{listing}{Listing}
\title{ Continual Learning with Scaled Gradient Projection }
\author {
    Gobinda Saha\textsuperscript{\rm 1},
    Kaushik Roy\textsuperscript{\rm 1}
}
\affiliations {
    \textsuperscript{\rm 1} Elmore Family School of Electrical and Computer Engineering\\
    Purdue University, West Lafayette, Indiana, USA\\
    gsaha@purdue.edu, kaushik@purdue.edu
}

\usepackage{bibentry}
\usepackage{hyperref}

\begin{document}

\maketitle

\begin{abstract}

In neural networks, continual learning results in gradient interference among sequential tasks, leading to catastrophic forgetting of old tasks while learning new ones. This issue is addressed in recent methods by storing the important gradient spaces for old tasks and updating the model orthogonally during new tasks. However, such restrictive orthogonal gradient updates hamper the learning capability of the new tasks resulting in sub-optimal performance. To improve new learning while minimizing forgetting, in this paper we propose a Scaled Gradient Projection (SGP) method, where we combine the orthogonal gradient projections with scaled gradient steps along the important gradient spaces for the past tasks. The degree of gradient scaling along these spaces depends on the importance of the bases spanning them. We propose an efficient method for computing and accumulating importance of these bases using the singular value decomposition of the input representations for each task. We conduct extensive experiments ranging from continual image classification to reinforcement learning tasks and report better performance with less training overhead than the state-of-the-art approaches. Codes: \href{https://github.com/sahagobinda/SGP}{\textcolor{magenta}{\texttt{https://github.com/sahagobinda/SGP}}}.  
\end{abstract}

\vspace{-10pt}
\section{Introduction}
Continual learning (CL)~\cite{cl1,cl2} aims to endue artificial intelligent (AI) systems with human-like adaptation capability in dynamically changing environments. In this learning paradigm, an AI model, commonly a deep neural network (DNN), learns from a sequence of tasks over time with the aim of accumulating and maintaining past knowledge and transferring it to future tasks. This objective is hard to achieve since standard optimization methods for training DNNs overwrite the parametric representations of past tasks with new input representations during model update~\cite{dynamic}. This leads to `Catastrophic Forgetting'~\cite{cat1,cat2} where performance of the past tasks degrades drastically, making continual learning a challenging problem. 

Many continual learning approaches for fixed capacity DNNs have been proposed that aim to balance two competing objectives: maintaining stability of past knowledge while providing sufficient plasticity for new learning. One line of works~\cite{ewc,oewc,ucb,pca_cl,ncl} achieves this goal by penalizing or preventing changes to the most important weights of the model for the past tasks while learning new tasks. Other works minimize forgetting by either storing samples from old tasks in the memory~\cite{Robbo,gem,er,hal,epr} or synthesizing old data by generative models~\cite{dgr} for rehearsal. Despite varying degrees of success, the stability-plasticity balance in such methods breaks down under long sequence of learning. Recently, gradient projection methods (Zeng et al.~\citeyear{owm}; Farajtabar et al.~\citeyear{ogd}; Saha et al.~\citeyear{gpm}; Wang et al.~\citeyear{adam-nscl}; Guo et al.~\citeyear{aop}) have shown superior performance to the other approaches. These methods minimize forgetting by minimizing gradient interference among tasks where new tasks are learned in the orthogonal gradient directions or spaces of the old tasks. Despite their remarkable stability (nearly zero forgetting) on the old tasks, performance of new tasks may suffer due to such restrictive gradient updates. To facilitate new learning, orthogonal gradient update is complemented with task similarity-based weight scaling~\citep{trgp} and dynamic gradient scaling~\citep{fs-dgpm}. However, these approaches either add significant training overheads or require old data storage, raising privacy concerns.   


\begin{figure*}[!h]
\begin{centering}
  \includegraphics[width=0.85\textwidth,keepaspectratio,page=2]{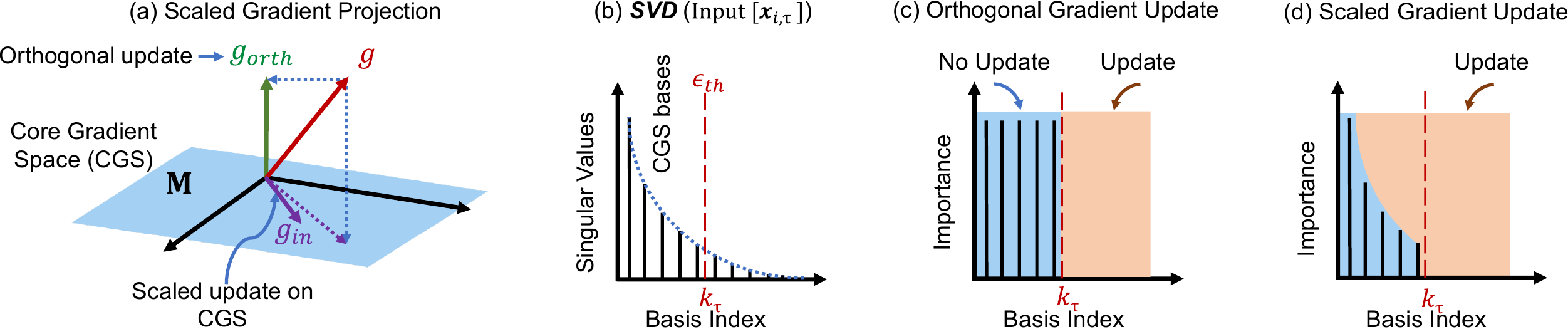}
  \caption{Scaled Gradient Projection (SGP) framework, (a) where new task is learned both in the orthogonal direction ($g_{orth}$) and along ($g_{in}$) the core gradient spaces (CGS) of the past tasks. (b) Bases of the CGS and importance of each basis are obtained from SVD on input representations. Unlike (c) orthogonal gradient projection where no update is allowed along the CGS, (d) in SGP, based on the importance of CGS bases, scaled gradient updates are made along those directions enabling better learning.} 
\label{fig:fig0}
\end{centering}
\vspace{-15pt}
\end{figure*} 

In this paper, we propose a new method - Scaled Gradient Projection (SGP) for continual learning that improves performance of the new task, while minimizing forgetting of the past tasks. Utilizing the framework introduced by Saha et al.~\citeyearpar{gpm}, we find the bases of the core (important) gradient spaces (CGS) for the past tasks by Singular Value Decomposition (SVD) of the input activations and store them in the memory. From the singular value distributions, we propose an efficient method for finding the (relative) importance of each basis of CGS. For each new task, we decompose the gradients into two components (Figure~\ref{fig:fig0}(a)) - one is along the past CGS and another is in the orthogonal direction to the CGS. In our method, the orthogonal gradient component is kept without modification and combined with a scaled version of the gradient component along the CGS. The degree of gradient scaling along each CGS basis is determined by importance of each basis stored in the memory. With this resultant gradient, the model is updated to achieve the CL objective. We evaluate our method in directly comparable settings on image classification datasets (CIFAR-100 and miniImageNet) and reinforcement learning tasks (consisting of six Atari games) and report better performance than the relevant baselines. We summarize our contributions as follows:  
\begin{itemize}
    \item We present Scaled Gradient Projection (SGP) for continual learning which, in addition to orthogonal gradient updates, allows scaled updates along the important gradient spaces of the past tasks, thus improving new learning. 
    \item We propose a method for finding and accumulating importance of each basis of those gradient spaces using singular values from the SVD on input representations. Our method does not store or use data from old tasks and has very little computing overhead.
    \item We evaluate SGP in diverse application domains: continual image classification and reinforcement learning tasks and obtain better performance than state-of-the-arts. We show, better task-level generalization and minimum forgetting enable superior performance in SGP.  
\end{itemize}

\section{Related Works}
Continual learning approaches can be broadly divided into three categories~\cite{cl3}.  

\noindent\textbf{Expansion-based Methods}: These methods prevent forgetting by expanding the network capacity. Notably, Progressive Neural Network (PNN)~\cite{pnn} adds new sub-network for each task and freezes them for next tasks. RCL~\cite{rcl} and BNS~\cite{BNS} perform network expansion with reinforcement learning, whereas Dynamic Expansion Network (DEN)~\cite{den} uses both model compression and expansion. APD~\cite{apd} minimizes such network growth by additive decomposition of parameters into task-specific and task-shared parts. In contrast, we learn continually in a fixed capacity model.   

\noindent\textbf{Regularization-based Methods}: These methods penalize changes in network parameters when learning new tasks based on their importance for prior tasks. Notably, Elastic Weight Consolidation (EWC)~\cite{ewc} computes this importance from the Fisher diagonal matrix after training, whereas Zenke et al.~\citeyearpar{si} find them during training from the loss sensitivity of the parameters. Other methods completely prevent updates to these important parameters by masking. For instance, PackNet~\cite{packnet} uses weight-level binary masks and HAT~\cite{hat} learns attention masks for neurons to prevent gradient updates. BLIP~\cite{blip} uses weight quantization and protects past information by bit-level freezing.    


\noindent\textbf{Memory-based Methods}: Memory-based methods can be divided into two sub-categories - (a) Experience Replay and (b) Gradient Projection based methods. In \text{experience replay} (ER)~\cite{icarl,er} data samples from the past tasks are stored in the memory and during new tasks, the model is jointly optimized with new data and memory data to mitigate forgetting. Meta Experience Reply~\cite{mer} and Look-ahead MAML~(Gupta et al.~\citeyear{lamaml}) combine experience replay with meta learning to achieve maximum knowledge transfer with minimum forgetting, whereas Dark Experience Replay~\cite{der} augments memory replay with logit distillation loss to improve performance. Alternatively, Gradient Episodic Memory (GEM)~\cite{gem} and Averaged-GEM (A-GEM)~\cite{agem} use memory data to compute gradient constraints for the new task so that loss on the past task does not increase. 

\text{Gradient projection} methods, in contrast, keep memory of gradient information from the past tasks and update model in the orthogonal directions of the past tasks to minimize forgetting. For instance, Orthogonal Weight Modulation (OWM)~\cite{owm} updates model in the orthogonal input directions of the old tasks, whereas~\citet{adam-nscl} performs model update in the null-space of the previous tasks. Orthogonal Gradient Descent~\cite{ogd} stores gradient directions of past tasks in the memory and project new gradient orthogonally to these memory gradients. Saha et al.~(\citeyear{gpm}) reduced training complexity and memory overhead of such methods in Gradient Projection Memory (GPM), where important gradient spaces for the past tasks are stored and new learning progresses in the orthogonal directions to these spaces. To relax orthogonal gradient constraint and encourage knowledge transfer to the new tasks, Trust Region Gradient Projection (TRGP)~\cite{trgp} augmented GPM with task-correlation based scaling of the weights of the old tasks. To achieve similar objective,~\citet{fs-dgpm} in FS-DGPM combined experience replay with gradient projection to dynamically scale the gradients. In a similar vein, we propose to improve new learning by scaled gradient projection. In contrast to previous methods, we do not learn these scales during training. Rather, we find and accumulate these scales in a one-shot manner after learning each task without using any old data. Thus we achieve better performance with significantly less training overhead while preserving data privacy. 

\section{Method}\label{sec:sgp_method}
In this section, we describe the steps for continual learning with our Scaled Gradient Projection (SGP) method. We consider supervised learning setup (Saha et al.~\citeyear{gpm}) where $T$ tasks are sequentially learned. Each task has a task descriptor, $\tau \in \{1,2....,T\}$ with a dataset, $\mathbb{D}_{\tau} = \{(\bm{x}_{i,\tau}, \bm{y}_{i,\tau})_{i=1}^{n_{\tau}}\}$ containing $n_{\tau}$ example pairs. We train a neural network having parameter set, $\mathbb{W}_\tau =\{(\bm{W}_\tau^l)_{l=1}^{L}\}$, where $L$ is the number of layers.  For each task we follow a three step process - (1) \textbf{training} the model on current task, (2) \textbf{finding bases} of the important gradient spaces for that task, and (3) \textbf{computing importance} of individual basis. These bases and corresponding importance are computed for each layer of the network and stored in the memory. The method described below is for a typical layer of a network and is generally applicable to all the layers. Hence we drop the layer notation. We use $\bm{x}_{i,\tau}$ to denote input or corresponding representation (activation) of input inside DNN for task $\tau$ interchangeably.   

\subsection{Learning Task, $\tau=1$}
\textbf{Train}: Starting from random initialization ($\mathbb{W}_0$), we train the model on dataset, $\mathbb{D}_1$ and obtain learned parameter set $\mathbb{W}_1$. During training on the first task no gradient scaling or constraint is imposed. 

\noindent\textbf{Find Important Gradient Space}: Next, we find important gradient spaces for this task using SVD on the input representations as described in GPM. For that, we randomly select $n_s$ training samples from $\mathbb{D}_1$ and pass it through the network. Then, for each layer we construct a representation matrix, $\bm{R}_{1} =[\bm{x}_{1,1}, \bm{x}_{2,1}, ..., \bm{x}_{n_s,1} ]$ by concatenating the corresponding input representations. To find the top-$k_1$ important basis of this input representation, SVD is performed on $\bm{R}_1 =\bm{U}_1\bm{\Sigma}_1\bm{V}_1^T$, where $\bm{U}_1$ and $\bm{V}_1$ are orthonormal matrices, and $\bm{\Sigma}_1$ has sorted singular values ($\sigma_{i,1}$) along its diagonal. This step is followed by $k_1$-rank approximation $(\bm{R}_1)_{k_1}$ of $\bm{R}_1$ using the norm-based criteria :
\begin{equation}\label{eq:gpm_eq6}
    \lvert\lvert(\bm{R}_1)_{k_1}\rvert\rvert_F^2 \geq \epsilon_{th}\lvert\lvert\bm{R}_1\rvert\rvert_F^2.
\end{equation}           
Here, $\lvert\lvert.\rvert\rvert_F^2$ is the Frobenius norm of the matrix. A brief description of SVD and $k$-rank approximation is given in Appendix~\ref{App_svd}. Threshold hyperparameter, $\epsilon_{th}\in (0,1)$ controls the value of $k_1$ selected. The first $k_1$ (column) vectors in $\bm{U}_1$ spans the most important space of inputs (representations) for task 1. Saha et al.~(\citeyear{gpm}) showed that these bases equivalently span the most important gradient space for task 1. We store these bases in $\bm{M} =[\bm{u}_{1,1}, \bm{u}_{2,1},..., \bm{u}_{k_1,1}]$, which is denoted as Gradient Projection Memory (GPM).    

\noindent\textbf{Compute Importance of Gradient Bases}: Typical gradient projection methods~(e.g. Saha et al.~\citeyear{gpm}) assume all the bases in GPM are equally important, hence no gradient step is taken along these directions for the new tasks to prevent forgetting. We find this assumption too restrictive for new learning and propose to assign variable importance to different bases. Our motivation comes from the observation that singular values ($\sigma_{i,1}$ ) corresponding to these bases ($\bm{u}_{i,1}$) have non-uniform distribution (Figure~\ref{fig:fig0}(b)) that implies their non-uniform importance for input or gradient preservation. Leveraging this distribution, we assign importance to individual basis. For that, we first get the singular value vector,  $\bm{\sigma_1}=\texttt{diag}(\bm{\Sigma_1},k_1)$, where $\bm{\sigma_1} \in \mathbb{R}^{k_1\times 1}$, corresponding to the top-$k_1$ basis in $\bm{U}_1$. We then compute importance for $i^{th}$ basis, where $i\in[1,k_1]$, by following:
\begin{equation}\label{eq:eq_scale}
    \lambda_{i,1} =\frac{(\alpha+1)\sigma_{i,1}}{\alpha\sigma_{i,1}+\texttt{max}(\bm{\sigma_1})}, 
\end{equation}
where $\alpha$ is a non-negative scale coefficient hyperparameter. The value of $\lambda_{i,1}$ will range from 0 to 1 as we are concerned with the non-negative singular values. We construct importance vector for this task by $\bm{\lambda}_1=[\lambda_{1,1},...,\lambda_{k_1,1}]^T$. Equation~\ref{eq:eq_scale} ensures that maximum importance (of $1$) is assigned to the basis with highest singular value and other bases are given importance ($<1$) relative to this maximum. In our formulation, $\lambda_{i,1}=1$ means no gradient step is allowed along the corresponding basis direction for the new tasks whereas along other bases gradients are scaled by the factor of ($1-\lambda_{i,1}$). In GPM~(Saha et al.~\citeyearpar{gpm}), all the bases in the memory are assumed to have maximum importance (of 1). This prevents any update along those directions (Figure~\ref{fig:fig0}(c)). In contrast, we allow a scaled gradient update along those bases (Figure~\ref{fig:fig0}(d)) enabling higher plasticity for new tasks, while importance based scaling ensures desirable stability of past tasks. Moreover, setting $\alpha$ to very high value (ideally $\infty$) in Equation~\ref{eq:eq_scale} would push all the importance values to 1. Thus for a large value of $\alpha$ our method converges to GPM. We transfer the basis matrix, $\bm{M}$ and accumulated importance vector $\bm{\lambda}$ to the next task. Here, $\bm{\lambda}=\bm{\lambda}_1$, $\bm{\lambda}\in \mathbb{R}^{k\times 1}$ with $k=k_1$ is the number of bases after task 1.

\subsection{Learning Task, $\tau\in[2,T]$}
\noindent\textbf{Train}: We learn the $\tau^{th}$ task sequentially using dataset, $\mathbb{D}_\tau$ only. Let $L_\tau$ denote loss for $\tau^{th}$ task. To prevent catastrophic forgetting and facilitate new learning we perform scaled gradient projection of new gradient, $\nabla_{\bm{W}_\tau} L_\tau$ as follows:
\begin{equation}\label{eq:sgp}
    \nabla_{\bm{W}_\tau} L_\tau = \nabla_{\bm{W}_\tau} L_\tau - (\bm{M} \bm{\Lambda}\bm{M}^T)(\nabla_{\bm{W}_\tau} L_\tau), 
\end{equation}
where importance matrix, $\bm{\Lambda}$ is a diagonal matrix of size $\mathbb{R}^{k\times k}$ containing $\bm{\lambda}$ in its diagonal. Here $k$ is the number of bases in $\bm{M}$ till task $\tau-1$. Such gradient projection ensures (a) the gradient components along orthogonal directions to $\bm{M}$ (GPM) will not be changed (as in Saha et al.~(\citeyear{gpm})), but (b) along the GPM bases the gradient components will be scaled by $(1-\lambda_i)$ based on the stored importance in $\bm{\lambda}$.   

\noindent\textbf{Update Important Gradient Space}: At the end of task $\tau$, we update $\bm{M}$ by adding the important gradient space for this task. For that, we construct $\bm{R}_\tau =[\bm{x}_{1,\tau}, \bm{x}_{2,\tau}, ..., \bm{x}_{n_s,\tau} ]$ using samples from $\mathbb{D}_\tau$ only. Now, SVD on $\bm{R}_\tau$ may produce some basis directions that are already present in $\bm{M}$ or are linear combinations of the bases in $\bm{M}$. These redundant bases are removed with the following step:
\begin{equation}\label{eq:gpm_eq9}
    \hat{\bm{R}}_{\tau} = \bm{R}_{\tau} - (\bm{M}\bm{M}^T)\bm{R}_{\tau} = \bm{R}_{\tau} -\bm{R}_{\tau,\bm{M}}. 
\end{equation}
Next, SVD is performed on $\hat{\bm{R}}_{\tau}=\hat{\bm{U}}_\tau\hat{\bm{\Sigma}}_\tau\hat{\bm{V}}_\tau^T$ and new $k_\tau$ bases are chosen for minimum $k_\tau$ satisfying the criteria:
\begin{equation}\label{eq:gpm_eq10}
    \lvert\lvert\bm{R}_{\tau,\bm{M}}\rvert\rvert_F^2+\lvert\lvert(\hat{\bm{R}}_\tau)_{k_\tau}\rvert\rvert_F^2 \geq \epsilon_{th}\lvert\lvert\bm{R}_\tau\rvert\rvert_F^2.
\end{equation}
Gradient space in GPM is updated (after $\bm{\lambda}$ update) by adding these new bases to $\bm{M}$ as $\bm{M} =[ \bm{M}, \hat{\bm{u}}_{1,\tau}, ..., \hat{\bm{u}}_{k_\tau,\tau} ]$.

\noindent\textbf{Update Importance of Gradient Bases}: This step involves assigning importance to the newly added $k_\tau$ bases and updating the importance of the old $k$ bases. Such importance update comes with several challenges. Firstly, SVD is performed on a part of the representation $\hat{\bm{R}}_{\tau}$, thus $\hat{\bm{\Sigma}}_\tau$ does not capture the full singular values distributions in $\bm{R}_{\tau}$. Hence, relative importance computation for the new $k_\tau$ bases using only the singular values of $\hat{\bm{\Sigma}}_\tau$ based on Equation~\ref{eq:eq_scale} would not be appropriate. Secondly, from Equation~\ref{eq:gpm_eq9} and Inequality~\ref{eq:gpm_eq10} we only know that old $k$ bases in $\bm{M}$ are responsible for capturing a fraction ($ \lvert\lvert\bm{R}_{\tau,\bm{M}}\rvert\rvert_F^2/\lvert\lvert\bm{R}_\tau\rvert\rvert_F^2$) of total norm of $\bm{R}_\tau$. However, computing their individual contributions in $ \lvert\lvert\bm{R}_{\tau,\bm{M}}\rvert\rvert_F^2 $, hence their importance for task $\tau$ is non-trivial. 

Here we propose a method for finding and updating the importance without using data from old tasks. First, we perform SVD on $\bm{R}_{\tau,M} =\bm{U}_{\tau,\bm{M}}\bm{\Sigma}_{\tau,\bm{M}}\bm{V}_{\tau,\bm{M}}^T $. By construction (Equation~\ref{eq:gpm_eq9}) the first $k$ (column) vectors in $\bm{U}_{\tau,\bm{M}}$ will be linear combinations of the bases in $\bm{M}$. Interestingly, SVD provides singular value vector, $\bm{\sigma}_{\tau,\bm{M}}=\texttt{diag}(\bm{\Sigma}_{\tau,\bm{M}},k)$ where each element signifies the importance of corresponding bases in $\bm{U}_{\tau,\bm{M}}$ for the current task $\tau$. Since these are redundant bases and will not be stored in $\bm{M}$, we transfer their singular values (hence importance) to the bases of $\bm{M}$. For that we create following projection coefficient matrix:
\begin{equation}\label{eq:sgp_c}
    \bm{C} =\bm{M^T}\bm{U}_{\tau,\bm{M}},  
\end{equation}
where each element $c_{i,j}$ is the dot product between $i^{th}$ columns of $\bm{M}$ and $j^{th}$ columns of $\bm{U}_{\tau,\bm{M}}$. Then we compute `surrogate singular values' for the bases in $\bm{M}$ by:
\begin{equation}\label{eq:sgp_sigma}
    \bm{\sigma}_{\tau,\bm{M}}^{'} =\sqrt{(\bm{C}\odot\bm{C})( \bm{\sigma}_{\tau,\bm{M}})^2}. 
\end{equation}
Here $\odot$ denotes element-wise multiplication, $(.)^2$ and $\sqrt{(.)}$ denote element-wise square and square root operations respectively. Next, we obtain the singular value vector, $\bm{\hat{\sigma}}_\tau=\texttt{diag}(\bm{\hat{\Sigma}}_\tau,k_\tau) \in \mathbb{R}^{k_\tau\times 1}$, for the newly added $k_\tau$ bases for the current task from SVD of $\bm{\hat{R}_\tau}$. Finally, we create the full ($k_\tau$-rank approximated) singular value vector, $\bm{\sigma}_\tau$ as: 

\begin{equation}\label{eq:sgp_singvec}
\bm{\sigma}_\tau = 
\begin{bmatrix}
\bm{\sigma}_{\tau,\bm{M}}^{'} \\
\bm{\hat{\sigma}}_\tau  \\
\end{bmatrix}
\in \mathbb{R}^{(k+k_\tau)\times 1}.  
\end{equation}
The above construction of $\bm{\sigma}_\tau$ ensures that the Inequality~\ref{eq:gpm_eq10} is satisfied. That can be re-written as:
\begin{equation}
    \sum_{i=1}^{k} (\sigma_{i,\tau,\bm{M}}^{'})^2 + \sum_{i=1}^{k_\tau} (\hat{\sigma}_{i,\tau})^2 \geq \epsilon_{th}\lvert\lvert\bm{R}_\tau\rvert\rvert_F^2.
\end{equation}
Therefore, we can use $\bm{\sigma}_\tau$ to obtain the basis importance vector, $\bm{\lambda}_\tau =f(\bm{\sigma}_\tau,\alpha)$ from Equation~\ref{eq:eq_scale} for the given $\tau^{th}$ task. Finally, we compute the accumulated importance over tasks and update the accumulated importance vector $\bm{\lambda}$. For $i\in[1,k]$ we update importance of old $k$ bases by: 

\begin{equation}\label{eq:eq_iacc}
        \lambda_i= 
\begin{cases}
    1,& \text{if } (\lambda_i+ \lambda_{i,\tau})\geq 1\\
     \lambda_i+ \lambda_{i,\tau},              & \text{otherwise}
\end{cases}
\end{equation}
We then add the importance of new $k_\tau$ bases in $\bm{\lambda}$ as $\bm{\lambda}=[\bm{\lambda}^T, \lambda_{k+1,\tau}, ... ,\lambda_{k+k_\tau,\tau}]^T$. After task $\tau$, the updated number of bases is $k=k+k_\tau$. We transfer $\bm{M}$ and $\bm{\lambda}$ to the next task and repeat the same procedure. The pseudocode of the algorithm is given in Algorithm~\ref{alg:sgp_algo} in the Appendix~\ref{App_algo}.  

\begin{figure*}[t]
\begin{centering}
  \includegraphics[width=0.95\textwidth,keepaspectratio,page=3]{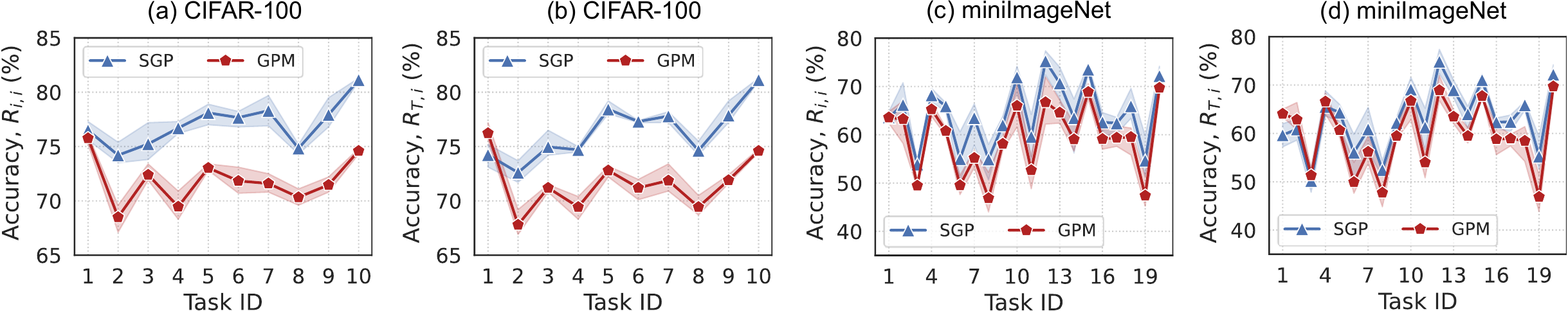}
  \caption{Test accuracy of each Split CIFAR-100 task (a) after learning that task, (b) after learning all the tasks. Test accuracy of each Split miniImageNet task (c) after learning that task, (d) after learning all the tasks.} 
\label{fig:fig1}
\end{centering}
\vspace{-5pt}
\end{figure*}

\section{Continual Image Classification Tasks}
\subsection{Experimental Setup}
\textbf{Datasets and Models}: First, we evaluate our method on standard image classification benchmarks~(Saha et al.~\citeyear{gpm}; Lin et al.~\citeyear{trgp}) including \textbf{Split CIFAR-100}, \textbf{CIFAR-100 Superclass} and \textbf{Split miniImageNet} datasets. The Split CIFAR-100 is constructed by dividing 100 classes of CIFAR-100~\cite{cifar} into 10 tasks with 10 distinct classes per task. CIFAR-100 Superclass~\cite{apd} is divided into 20 tasks where each task contains 5 semantically related classes from CIFAR-100 dataset, whereas Split miniImageNet is constructed by dividing 100 classes of miniImageNet~\cite{miniI} into 20 tasks with 5 distinct classes per task. We do not use any data augmentation in our experiments. Similar to Saha et al.~\citeyearpar{gpm}, we use 5-layer AlexNet for Split CIFAR-100, LeNet for CIFAR-100 Superclass, and reduced ResNet18 for Split miniImageNet experiments. We train and test all the methods in task-incremental learning~\cite{eval1} setup where each task has a separate classifier head. Details of dataset statistics and network architectures are given in Appendix~\ref{App_exp_details}.  

\noindent\textbf{Baselines}: As baselines for comparison, we use EWC~\cite{ewc} and HAT~\cite{hat} from the regularization-based methods, OWM~\cite{owm}, GPM~(Saha et al.~\citeyear{gpm}), TRGP~\cite{trgp} and FS-DGPM~\cite{fs-dgpm} from gradient projection methods, and Experience Replay with reservoir sampling (ER\_Res)~\cite{er} and  A-GEM~\cite{agem} from replay methods. To provide an upper limit for performance, we include an oracle baseline - Multitask where all the tasks are jointly learned in a single network using the entire dataset. Additionally, for CIFAR-100 Superclass we provide comparisons with PNN, RCL, DEN, and APD~\cite{apd} from the expansion-based methods.  

\noindent\textbf{Training Details}: All the baselines and our method are trained with plain stochastic gradient descent (SGD). Following GPM setup, each task in Split CIFAR-100 and CIFAR-100 Superclass is trained for 200 and 50 epochs respectively with early stopping criteria with batch size of 64, whereas each Split miniImageNet task is trained for 10 epochs with batch size of 10. We use the same threshold values ($\epsilon_{th}$) as GPM and use scale coefficient ($\alpha$) of 1, 3 and 10 for Split miniImageNet, CIFAR-100 Superclass and Split CIFAR-100 datasets respectively. More details of training setup, hyperparameters considered in baselines and our methods and implementation are given in Appendix~\ref{App_exp_details}.  

\noindent\textbf{Evaluation Metrics}: For evaluations, we use two performance metrics~\cite{gem} : (1) \textbf{ACC} - measures average test classification accuracy of all tasks and (2) \textbf{BWT} (backward transfer) - measures influence of new learning on the past knowledge where negative BWT implies forgetting. ACC and BWT are defined as: 
\begin{equation}\label{eq:sgp_acc}
    \text{ACC} = \frac{1}{T}\sum_{i=1}^T R_{T,i};\text{BWT} = \frac{1}{T-1}\sum_{i=1}^{T-1} R_{T,i} - R_{i,i}. 
\end{equation}
Here, $T$ is the total number of tasks. $R_{T,i}$ is the accuracy of the model on $i^{th}$ task after learning the $T^{th}$ task sequentially.

\renewcommand{\arraystretch}{1.3}
\begin{table}[t]\centering
\hspace{5cm}
\scalebox{0.83}{
\begin{tabular}{@{}lcccc@{}}\toprule\label{tab_app_4}
&  \multicolumn{2}{c}{\textbf{CIFAR-100}} &  \multicolumn{2}{c}{\textbf{miniImageNet}} \\
\cmidrule{2-3} \cmidrule{4-5} 
\textbf{Methods}& {ACC (\%)}  & {BWT}  & {ACC (\%)}  & {BWT}\\ 
\midrule
Multitask\textsuperscript{$\dagger$} & 79.58 \scriptsize $\pm$ 0.54 & - &  69.46 \scriptsize$\pm$ 0.62 & - \\
\midrule
OWM\textsuperscript{$\dagger$}   &  50.94 \scriptsize $\pm$ 0.60 & - 0.30 \scriptsize $\pm$ 0.01  & - & -  \\
A-GEM\textsuperscript{$\dagger$}   &  63.98 \scriptsize $\pm$ 1.22 & - 0.15 \scriptsize $\pm$ 0.02 &  57.24 \scriptsize$\pm$ 0.72 & - 0.12 \scriptsize$\pm$ 0.01\\
ER\_Res\textsuperscript{$\dagger$}&  71.73 \scriptsize $\pm$ 0.63 & - 0.06 \scriptsize $\pm$ 0.01  &  58.94 \scriptsize$\pm$ 0.85 & - 0.07 \scriptsize$\pm$ 0.01 \\
EWC\textsuperscript{$\dagger$}   &  68.80 \scriptsize $\pm$ 0.88 & - 0.02 \scriptsize $\pm$ 0.01 &  52.01 \scriptsize$\pm$ 2.53 & - 0.12 \scriptsize$\pm$ 0.03 \\
HAT\textsuperscript{$\dagger$}   &  72.06 \scriptsize $\pm$ 0.50 & - 0.00 \scriptsize $\pm$ 0.00 &  59.78 \scriptsize$\pm$ 0.57 & - 0.03 \scriptsize$\pm$ 0.00 \\
FS-DGPM\textsuperscript{$\ddagger$} &  74.33 \scriptsize $\pm$ 0.31 &  - 0.03 \scriptsize $\pm$ 0.00 & - & - \\
TRGP &  74.46 \scriptsize $\pm$ 0.32\textsuperscript{*} &  - 0.01 \scriptsize $\pm$ 0.00\textsuperscript{*} & 60.93 \scriptsize$\pm$ 0.94 & -0.00 \scriptsize$\pm$ 0.00 \\
GPM\textsuperscript{$\dagger$} &  72.48 \scriptsize $\pm$ 0.40 &  - 0.00 \scriptsize $\pm$ 0.00 & 60.41 \scriptsize$\pm$ 0.61 & -0.00 \scriptsize$\pm$ 0.00 \\
\midrule
SGP (ours) &  \textbf{76.05 \scriptsize $\pm$ 0.43} &  - 0.01 \scriptsize $\pm$ 0.00 & \textbf{62.83 \scriptsize$\pm$ 0.33} & -0.01 \scriptsize$\pm$ 0.01 \\
\bottomrule
\end{tabular}}
\caption{Results (mean $\pm$ std in 5 runs) on Split CIFAR-100 and Split miniImageNet. \textsuperscript{$\dagger$}, \textsuperscript{$\ddagger$} and \textsuperscript{*} denote the results from GPM, FS-DGPM and TRGP respectively.}
\label{tab:tab1} 
\vspace{-10pt}
\end{table}
\subsection{Results and Discussions}
\textbf{Performance Comparison}: First, we compare the ACC and BWT of our method with the baselines. Table~\ref{tab:tab1} shows the comparisons for Split CIFAR-100 and Split miniImageNet dataset. In Split CIFAR-100 tasks, SGP obtains highest ACC ($76.05\pm0.43\%$) with accuracy gain of $\sim3.5\%$ over GPM and $\sim1.6\%$ over TRGP and FS-DGPM. Compared to the best performing regularization method HAT and replay method ER\_Res, SGP obtains $\sim4\%$ improvement in accuracy. In terms of BWT, SGP shows $\sim1\%$ forgetting which is lower than FS-DGPM, comparable to TRGP and slightly higher than GPM and HAT. Similar trend is observed for longer (20) tasks sequence in miniImageNet dataset, where SGP obtains accuracy gain of $\sim2.4\%$ over GPM and $\sim2\%$ over TRGP with BWT of $-1.5\% $. Compared to GPM, this slight increase in forgetting (negative BWT) in SGP is intuitive, as accuracy gain in SGP comes from selective model updates along the important gradient spaces of the past tasks which interfere (minimally) with old learning. Finally, we compare the performance of gradient projection methods with the expansion-based methods in Table~\ref{tab:tab2} for 20-task CIFAR-100 Superclass dataset. Here, SGP achieves highest ACC ($59.05\pm0.50\%$) with $-1.5\%$ BWT obtaining accuracy improvement of $2.2\%$ over APD with $30\%$ fewer network parameters. It has $1.3\%$ ACC gain over GPM and performs marginally better than TRGP and FS-DGPM. Overall, improved performance in SGP indicates that scaled gradient projections offer better stability-plasticity balance in CL compared to strict orthogonal projections.

\renewcommand{\arraystretch}{1.1}
\begin{table*}[t]\centering

\centering
\vspace{-15pt}
\scalebox{0.85}{
\begin{tabular}{@{}lcccccccccc@{}}\label{tab_3}
&  \multicolumn{5}{c}{}  \\
\toprule
&  \multicolumn{9}{c}{\textbf{Methods}} \\ 
\cmidrule{2-11}
Metric &  \multicolumn{1}{c}{STL\textsuperscript{$\dagger$}} & \multicolumn{1}{c}{PNN\textsuperscript{$\dagger$}} & \multicolumn{1}{c}{DEN\textsuperscript{$\dagger$}}& \multicolumn{1}{c}{RCL\textsuperscript{$\dagger$}}& \multicolumn{1}{c}{APD\textsuperscript{$\dagger$}} & \multicolumn{1}{c}{EWC\textsuperscript{$\ddagger$}}  & \multicolumn{1}{c}{FS-DGPM\textsuperscript{$\ddagger$}} & \multicolumn{1}{c}{TRGP\textsuperscript{$\dagger$}} & \multicolumn{1}{c}{GPM\textsuperscript{$\dagger$}} & \multicolumn{1}{c}{\textbf{SGP (ours)}}\\
\midrule
ACC (\%)  & 61.00 &   50.76  &  51.10 &   51.99  & 56.81 & 50.26 & 58.81 & 58.25 & 57.72 & \textbf{59.05} \\
Capacity (\%)  & 2000 &   271  & 191&   184  & 130 & 100&  100 & 100 & 100& 100\\
\bottomrule
\end{tabular}}
\caption{\textcolor{black}{Results (averaged over 5 task-order) for CIFAR-100 Superclass. \textsuperscript{$\dagger$} and \textsuperscript{$\ddagger$}~denote the results from TRGP and FS-DGPM respectively. Single task learning (STL) is a non-continual learning baseline where each task is learned in a separate network.}}
\label{tab:tab2}
\end{table*}

\begin{figure*}[!h]
\begin{centering}
  \includegraphics[width=0.87\textwidth,keepaspectratio,page=4]{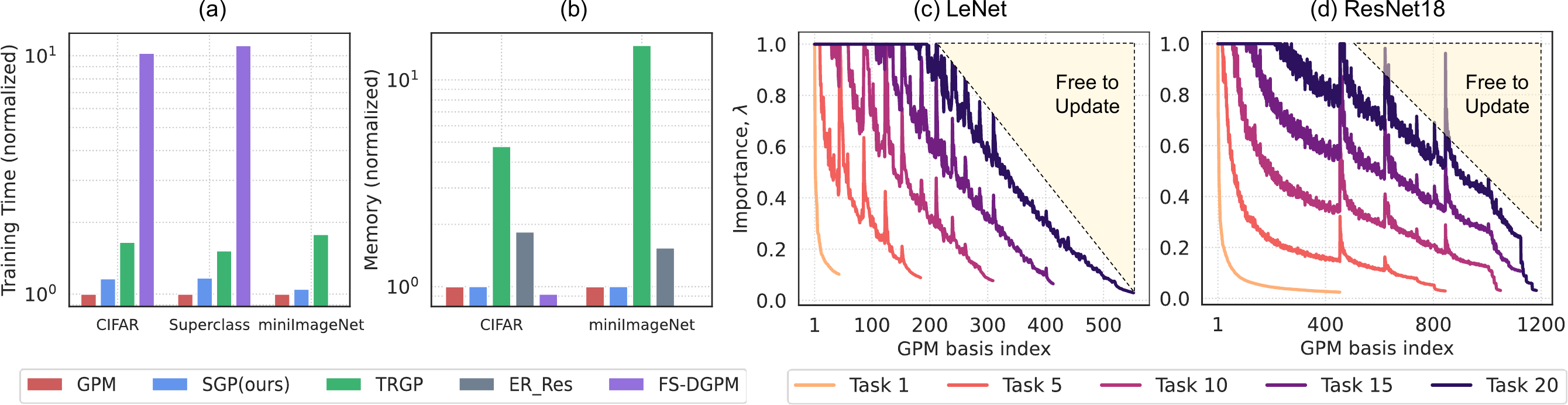}
  \caption{(a) Training time and (b) Memory comparisons (both normalized with respect to GPM). Smaller means better. Basis importance distribution at the last layer before classifier of the model in (c) CIFAR-100 Superclass and (d) miniImageNet tasks. } 
\label{fig:fig2}
\end{centering}
\end{figure*}

\begin{figure}[!h]
\begin{centering}
  \includegraphics[width=0.4\textwidth,keepaspectratio,page=5]{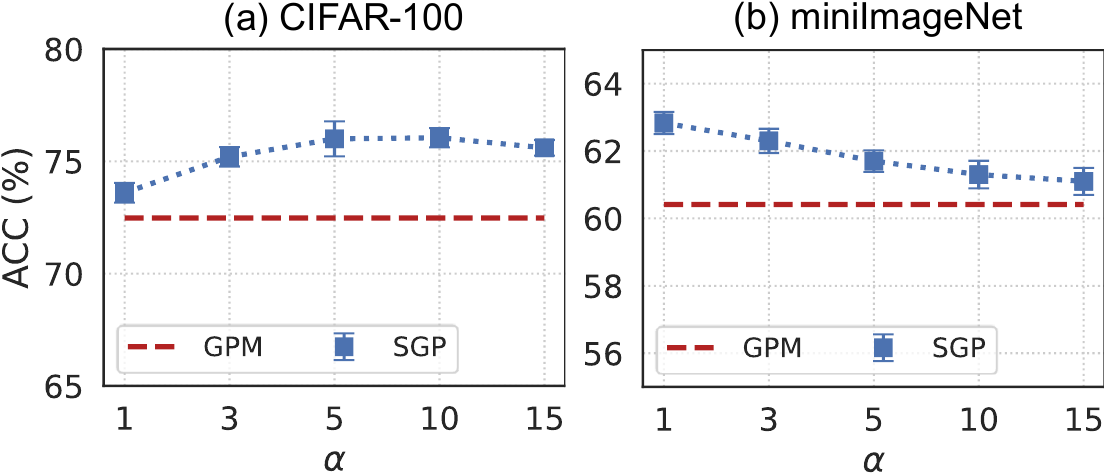}
  \caption{Impact of varying $\alpha$ on SGP performance.} 
\label{fig:fig3}
\end{centering}
\vspace{-15pt}
\end{figure} 
       
\noindent\textbf{Task-Level Generalization and Forward Transfer}: To quantify task-level generalization and forward transfer, in Figure~\ref{fig:fig1}(a) and (c) we plot $R_{i,i}$, which is the test accuracy of $i^{th}$ task immediately after learning $i^{th}$ task during CL process, for all Split CIFAR-100 and Split miniImageNet  tasks. Analysis for CIFAR-100 Superclass is given in the Appendix~\ref{App_add_results_1} (Figure~\ref{fig:app_fig1}). Compared to GPM, in SGP each new task ($\tau>1$) has better accuracy which indicates relaxing constraints with gradient scaling (along CGS) facilitates learning and improves generalization capability of the new tasks. Forward transfer (FWT)~(Veniat et al.~\citeyear{fwt}; Lin et al. \citeyear{trgp}) is quantified by the difference in accuracy when a task is learned sequentially and when that task is learned standalone in a single network. Positive FWT thus means sequential learning has improved the performance, hence a desired attribute of CL. Two CL methods (e.g. A \& B) can be compared in terms of relative FWT~\cite{trgp} as:
\begin{equation}
    \Delta_{A-B} = \frac{1}{T}\sum_{i=1}^{T} R_{i,i}^A - R_{i,i}^B.
\end{equation}
From Figure~\ref{fig:fig1}(a) and (c), we find that relative forward transfer between SGP and GPM, $\Delta_{SGP-GPM}$ is approximately $+5\%$ and $+4\%$ for Split CIFAR-100 and Split miniImageNet respectively. This results again show better task-level generalization is achieved in CL with SGP. 

In Figure~\ref{fig:fig1}(b) and (d) we plot $R_{T,i}$ - final test accuracy of each task after learning all $T$ tasks, for Split CIFAR-100 and miniImageNet. Comparing them to Figure~\ref{fig:fig1}(a) and (c), we find that there is a slight difference in final and initial task accuracies ($R_{T,i}-R_{i,i}$) in SGP which translates to backward transfer according to Equation~\ref{eq:sgp_acc}. Moreover, earlier tasks have a slightly higher drop, whereas later tasks show negligible to no drop in accuracy. Overall, almost all the tasks (except few earlier ones) have higher final accuracy in SGP than GPM which translates to better average accuracy (ACC). 

\noindent\textbf{Training Time and Memory Overhead}: In Figure~\ref{fig:fig2}(a) we compare the training complexity of gradient projection methods in terms of relative (wall-clock) time needed for sequentially training all the tasks. Here the normalization is performed with respect to GPM. Compared to GPM, SPG tasks only up to $17\%$ more time which is used for basis importance calculations. TRGP takes up to $78\%$ more time which includes trust regions selection and weight scale learning. Whereas, FS-DGPM takes an order of magnitude more time than GPM and SGP primarily due to iterative sharpness evaluation steps used for dynamic gradient projection, potentially limiting its application in deeper networks. More details of time measurements and comparison with other methods are provided in Appendix~\ref{App_add_results_1}. Next, in Figure~\ref{fig:fig2}(b), we show the extra memory used by different memory-based methods either for gradient or data storage. Compared to GPM, SGP takes only $0.1\%$ extra memory for storing basis importance. TRGP uses up to an order of magnitude more memory for storing weight scales for each layer for each task. Thus, SGP provides performance gain in CL with significantly less training time and memory overhead.   

\begin{figure*}[t]
\begin{centering}
  \includegraphics[width=0.91\textwidth,keepaspectratio,page=6]{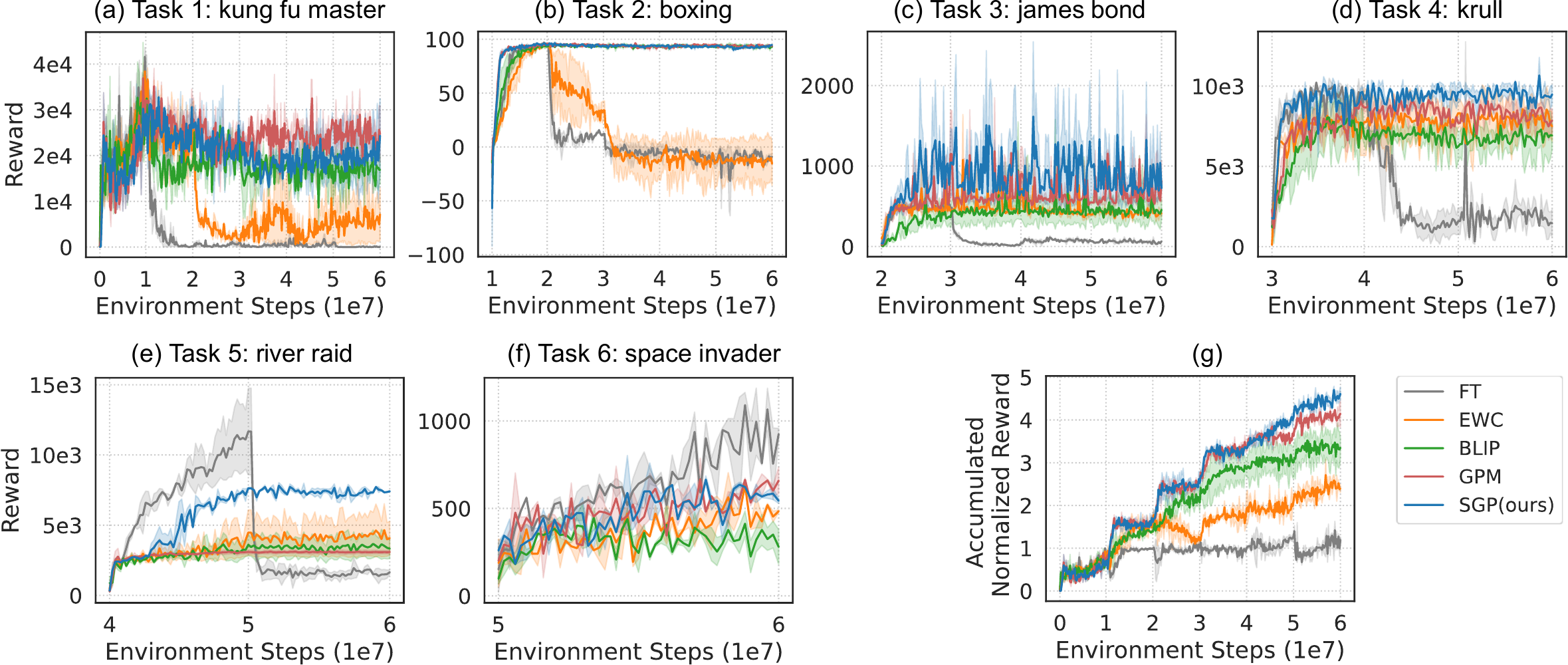}
  \caption{(a)-(f) Rewards on six sequential Atari games (tasks). Each task is trained for 10 million environment steps, and 60 million environment steps are used in total. Results are averaged over 3 random seeds. (g) Accumulated Normalized Rewards.} 
\label{fig:fig4}
\end{centering}
\vspace{-15pt}
\end{figure*}

\noindent\textbf{Importance Accumulation Dynamics}: Next, we analyze the basis importance accumulation dynamics across tasks in SGP. Figure~\ref{fig:fig2}(c) and (d) show importance distributions across different tasks at the last layer before the classifier of LeNet and ResNet18 in CIFAR-100 Superclass and miniImageNet experiments. These distributions reflect non-uniformity of the singular value distributions from which they are generated and help us understand the degree of gradient constraint in SGP optimization. For task 1, only one basis per layer has the importance of $\lambda_1=1$, whereas large number of bases (especially in Figure.~\ref{fig:fig2}(d)) have very low importance. Unlike GPM, where gradients along all these directions are blocked, in SGP large steps are encouraged along the less important basis directions relaxing the gradient constraints significantly. As new tasks are learned, new bases are added and importance of the old bases is updated (Equation~\ref{eq:eq_iacc}). Due to these accumulative updates, importance of the bases gradually moves towards unity making the optimization gradually restrictive along those directions to encounter forgetting. This process still leaves significant degree of freedom for future learning. As in the last layer of LeNet and ResNet18, even after learning 20 tasks, SGP fully restricts updates along only $\sim37\%$ and $\sim18\%$ of the stored basis directions, whereas along the remaining directions gradient steps with variable scaling are allowed. In Appendix~\ref{App_add_results_1} (Figure~\ref{fig:app_imp1} \&~\ref{fig:app_imp2}), importance ($\lambda$) distributions in all the layers (except classifier) for Split CIFAR-100 and CIFAR-100 Superclass are shown where a similar trend is observed.        

\noindent\textbf{Impact of Scale Coefficient ($\alpha$)}: In Figure~\ref{fig:fig3} we show the impact of varying $\alpha$ on SGP performance. In Figure~\ref{fig:fig2}(c), (d) we observe that the importance distributions vary with the network architecture and dataset. Thus depending on these factors optimum performance will occur at different $\alpha$. Ultimately, at large enough $\alpha$ the SGP performance will be similar to GPM. For Split CIFAR-100 we find that for $\alpha$ in the range of 5 to 10, SGP has maximum ACC. In miniImageNet at $\alpha=1$ best performance is obtained and increasing $\alpha$ pushes SGP performance towards GPM. CIFAR-100 Superclass results are given in Appendix~\ref{App_add_results_1} (Figure~\ref{fig:app_fig1}(c)).   
\section{Continual Reinforcement Learning Tasks}\label{clrl}
\subsection{Experimental Setup}
Unlike majority of CL works, we test the scalability of our algorithm in challenging yet highly relevant continual reinforcement learning benchmarks. For that, we follow the state-of-the-art setup in BLIP~\cite{blip}, where the agent sequentially learns to play the following six Atari~\cite{atari} games: \textbf{kung fu master, boxing, james bond, krull, river raid, space invaders}. We use a PPO agent~\cite{ppo} with 3 convolution layers (32-64-128 filters) followed by a fully connected layer (1024 neurons) that is trained with initial learning rate $2.5\times 10^{-4}$ and entropy regularization coefficient 0.01. We sample 10 million steps in each (task) environment to train our agents. We use EWC, BLIP, and naive baseline of sequential finetuning (FT) for comparison. Like baselines, we could not use standard \texttt{Adam} optimizer~\cite{adam} in GPM and SGP as it shows catastrophic forgetting when projected gradients are passed into \texttt{Adam} (see Appendix Figure~\ref{fig:app_adam_fail}). Hence we take the gradient output from \texttt{Adam} and then perform orthogonal or scaled projection on that (see modified \texttt{Adam} algorithm in Appendix~\ref{App_add_results_2}). For GPM and SGP, $\epsilon_{th}=0.995$ and for SGP $\alpha=25$ is used.  
\subsection{Results and Discussions} 
To evaluate and compare performance, following BLIP, we score each game (task) by average reward obtained after playing that game for 10 episodes. In Figure~\ref{fig:fig4}(a)-(f), we plot rewards of the PPO agents on each task from the point they start to train on the task to the end of the whole continual learning process. In each of these plots, for the first 10 million steps, the agent is trained on that task and at regular intervals within that period the agent's scores (reward) on that task are plotted. In the remaining period, the agent is not trained on that task, only its performance on that task is tested and plotted. From these plots, we observe that FT trained agent suffers from catastrophic forgetting as its reward on one task rapidly decreases after learning next task. EWC trained agent forgets first two tasks and maintains the performance of last four tasks. BLIP trained agent, on the other hand, maintains performance for each task, however from task 3 to 6, compared to SGP, it obtains significantly less rewards primarily due to capacity constraints. Similar trend is observed in GPM, especially in task 5 GPM underperforms SGP trained agent by a large margin due to strict gradient constraints. Overall, in all these tasks SGP has comparable or better (especially later tasks) performance than other methods taking advantage of scaled gradient updates.     
 
Finally, to compare the performance of each method in an overall manner, in Figure~\ref{fig:fig4}(g) we plot the accumulated reward, which is the sum of rewards on all learned tasks at each environment step. We normalize rewards for each task on a scale of 0 to 1. This means an agent's accumulated normalized reward should increase steadily from 0 to 6 if it learns and remembers all six tasks. In contrast, when an agent continually forgets past tasks, which is the case for FT in Figure~\ref{fig:fig4}, then the accumulated normalized reward should oscillate around 1. In this metric, SGP outperforms all the other methods in the entire continual learning sequence, particularly accumulating $12\%$ and $36\%$ more rewards compared to GPM and BLIP respectively. These experiments show that SGP can provide stability to past learning, while improving new learning by better utilizing gradient capacity in DNNs, even under long sequence of gradient updates.     
 
\section{Conclusions}
In summary, to improve continual learning performance, we propose a scaled gradient projection method that combines orthogonal gradient projections with a scaled gradient updates along the past important gradient spaces. We introduce a principled method for finding the importance of each basis spanning these spaces and use them to guide the gradient scaling. With quantitative analyses, we show that our method enables better task-level generalization with minimum forgetting by effectively utilizing the capacity of gradient spaces in the DNNs. Moreover, we show our method has very little training overhead and is scalable to CL applications in multiple domains including image classification and reinforcement learning. Future works should take complementary benefits of scaled gradient projection by combining it with other methods in various CL setups. 

\section*{Acknowledgements}
This work was supported in part
by the National Science Foundation, Vannevar Bush Faculty
Fellowship, Army Research Office, MURI, 
DARPA AI Exploration (ShELL),
and by Center
for Brain-Inspired Computing (C-BRIC), one of six centers
in JUMP, a SRC program sponsored by DARPA.
\bibliography{aaai23}


\clearpage

\appendix
\onecolumn
\section*{Appendix}
Section A provides a brief introduction to SVD and $k$-rank approximation. Pseudocode of the SGP algorithm is given in Section B. Experimental details including a list of hyperparameters and additional results and analyses for continual image classification tasks are provided in Section C and D respectively. Additional results and analyses for continual reinforcement learning tasks are provided in section E. 

\section{SVD and $k$-rank Approximation}\label{App_svd}

Singular Value Decomposition (SVD) can be used to factorize a rectangular matrix, $\bm{A} =\bm{U}\bm{\Sigma}\bm{V}^T \in \mathbb{R}^{m\times n}$ into the product of three matrices, where $\bm{U}\in \mathbb{R}^{m\times m}$ and $\bm{V}\in \mathbb{R}^{n\times n}$ are orthonomal matrices, and $\bm{\Sigma} $ is a diagonal matrix that contains the sorted singular values along its main diagonal (Deisenroth et al.~\citeyear{mml}). If the rank of the matrix is $r$ ($r\leq \text{min}(m,n)$), $\bm{A}$ can be expressed as $\bm{A}=\sum_{i=1}^r \sigma_i\bm{u}_i \bm{v}_i^T$, where $\bm{u}_i \in \bm{U}$ and $\bm{v}_i \in \bm{V}$ are left and right singular vectors and $\sigma_i \in diag(\bm{\Sigma})$ are singular values. \noindent$k$-rank approximation of $\bm{A}$ can be written as, $\bm{A}_k=\sum_{i=1}^k \sigma_i\bm{u}_i \bm{v}_i^T$, where $k\leq r$ and its value can be chosen by the smallest $k$ that satisfies the norm-based criteria : $\lvert\lvert \bm{A}_k \rvert\rvert_F^2 \geq \epsilon_{th}\lvert\lvert\bm{A}\rvert\rvert_F^2$. Here, $\lvert\lvert.\rvert\rvert_F$ denotes the Frobenius norm of the matrix and $\epsilon_{th} \in(0,1)$ is the threshold hyperparameter.

\section{SGP Algorithm Pseudocode}\label{App_algo}

\begin{algorithm}
   \caption{Algorithm for Continual Learning with SGP}
   \label{alg:sgp_algo}
    \begin{algorithmic}[1]
       \Procedure{TrainSGP}{$f_{\bm{W}}$, $\mathcal{D}^{train}$,$\eta$ ,$\epsilon_{th}$, $\alpha$}
       \State \textbf{Inputs}: $f_{\bm{W}}$: neural network, $\mathcal{D}^{train}$: training dataset, $\eta$: learning rate ,$\epsilon_{th}$: threshold, $\alpha$: scale coefficient
       \State Initialize, $\bm{M}^l \gets$ [\space] and $\bm{\lambda}^l \gets$ [\space], for all $l=\{1,2,....,L-1\} $      \Comment{L is the number of layers}   
       
       \State $\mathcal{M} \gets \{(\bm{M}^l)_{l=1}^{L-1}\}$ \Comment{$\bm{M}^l$ is the basis matrix (GPM) at layer $l$} 
       \State $\mathcal{S} \gets \{(\bm{\lambda}^l)_{l=1}^{L-1}\}$
       \Comment{$\bm{\lambda}^l$ is the accumulated importance vector at layer $l$} 
    
    \State $\bm{W} \gets \bm{W}_0$ \Comment{network parameter, $\bm{W}$ initialization }
      \For {$\tau \in {1,2,.....,T}$}    
      \Repeat
      \State \textit{$B_n$} $\overset{\mathrm{n}}{\sim} \mathcal{D}_\tau^{train}$ \Comment{sample a mini-batch of size $n$ from task $\tau$}
      \State gradient, $\nabla_{\bm{W}}L_{\tau} \gets \texttt{getgradient}(\textit{$B_n$}, f_{\bm{W}} )$ \Comment{compute gradient}
       \State $\nabla_{\bm{W}}L_{\tau} \gets \texttt{ScaledGradientProjection}(\nabla_{\bm{W}}L_{\tau},\mathcal{M},\mathcal{S} )$ \Comment{see Equation~\ref{eq:sgp}}
       \State $\bm{W} \gets \bm{W} - \eta \nabla_{\bm{W}}L_{\tau}$ \Comment{model update}
      \Until{convergence}
        \State
        \State // Update Basis matrix and Importance Vector 
      \State \textit{$B_{n_s}$} $\overset{\mathrm{n_s}}{\sim} \mathcal{D}_\tau^{train}$ \Comment{sample a mini-batch of size $n_s$ from task $\tau$}
      \State // Construct representation matrices for each layer by forward pass (see section~\ref{sec:sgp_method})
      \State $\mathcal{R}_\tau \gets \texttt{forward} (\textit{$B_{n_s}$}, f_{\bm{W}} )$, where $\mathcal{R_\tau} = \{(\bm{R}_\tau^l)_{l=1}^{L-1}\}$
      
        \For{layer, $l=1,2,...L-1$}
        \State $\bm{R}_{\tau,\bm{M}^l}^l \gets (\bm{M}^l(\bm{M}^l)^T)\bm{R}_\tau^l$
        \State $\hat{\bm{R}}_\tau^l \gets (\bm{R}_\tau^l -\bm{R}_{\tau,\bm{M}^l}^l )$ \Comment{see Equation (\ref{eq:gpm_eq9})}
        \State $\hat{\bm{U}}_\tau^l,\hat{\bm{\Sigma}}_\tau^l \gets \textbf{\texttt{SVD}} (\hat{\bm{R}}_\tau^l)$
        \State $k_\tau \gets \texttt{criteria} (\hat{\bm{R}}_\tau^l, \bm{R}_\tau^l, \epsilon_{th}^l) $ \Comment{see Inequality (\ref{eq:gpm_eq10})}
        \State $k \gets \bm{M}^l.\texttt{shape[1]}$ \Comment{get the number of (old) basis in $\bm{M}^l$ after task $\tau-1$}
        \State $\bm{U}_{\tau,\bm{M}^l}^l, \bm{\Sigma}_{\tau,\bm{M}^l}^l \gets \textbf{\texttt{SVD}} (\bm{R}_{\tau,\bm{M}^l}^l)$
        \State $\bm{C}^l \gets (\bm{M}^l)^T\bm{U}_{\tau,\bm{M}^l}^l$ \Comment{$\bm{U}_{\tau,\bm{M}^l}^l$ has the first $k$ columns}
        \State $\bm{\sigma}_\tau^l \gets \texttt{SingularValueVector}(\bm{C}^l,\bm{\Sigma}_{\tau,\bm{M}^l}^l,\hat{\bm{\Sigma}}_\tau^l)$ \Comment{see Equation~\ref{eq:sgp_sigma} \&~\ref{eq:sgp_singvec}}
        \State $\bm{\lambda}_\tau^l \gets f(\bm{\sigma}_\tau^l,\alpha)$ \Comment{importance vector for $\tau^{th}$ task at layer $l$, see Equation~\ref{eq:eq_scale}}
        \State $\bm{\lambda}^l \gets \texttt{AccumulateImportance}(\bm{\lambda}^l,\bm{\lambda}_\tau^l,k)$ \Comment{importance accumulation for the old $k$ basis (Equation~\ref{eq:eq_iacc})}
        \State $\bm{\lambda}^l \gets [(\bm{\lambda}^l)^T, (\bm{\lambda}_\tau^l[k:k+k_\tau])^T]^T$ \Comment{\textbf{Accumulated importance vector update}}
        \State $\bm{M}^l \gets [\bm{M}^l, \hat{\bm{U}}_\tau^l[0:k_\tau]]$ \Comment{\textbf{Basis matrix (GPM) update}}
        
        \EndFor
    \EndFor
    \State \textbf{return} $f_{\bm{W}}, \mathcal{M}, \mathcal{S}$ 
    \EndProcedure
    \end{algorithmic}
\end{algorithm}

\section{Experimental Details: Continual Image Classification Tasks}\label{App_exp_details}
\subsection{Dataset Statistics}
Table~\ref{tab:sgp_app_1} shows the summary of the datasets used in the experiments.  
\vspace{-5pt}

\renewcommand{\arraystretch}{1.1}
\begin{table*}[h]\centering
\small
\scalebox{0.88}{
\begin{tabular}{@{}lccccc@{}}\toprule
&  \multicolumn{1}{c}{\textbf{Split CIFAR-100}} & \phantom{a}& \multicolumn{1}{c}{\textbf{CIFAR-100 Superclass}} & \phantom{a}& \multicolumn{1}{c}{\textbf{Split miniImageNet}}\\
\midrule
num. of tasks    & 10 &&   20  && 20   \\
input size   & $3\times 32\times 32$ && $3\times 32\times 32$ && $3\times 84\times 84$\\
\# Classes/task  & 10 && 5 && 5  \\
\# Training samples/tasks  &4,750&&2,375&& 2,375\\
\# Validation Samples/tasks &250&& 125&&125\\
\# Test samples/tasks & 1,000&& 500&& 500\\
\bottomrule
\end{tabular}}
\caption{Dataset Statistics.}
\label{tab:sgp_app_1}
\end{table*}
\vspace{-10pt}

\begin{table*}[h]
\centering
\scalebox{0.88}{
\begin{tabular}{@{}lcl@{}}\toprule
\textbf{Methods}& \phantom{a} & \textbf{Hyperparameters} \\
\midrule
OWM    & & lr :  0.01 (cifar) \\
\midrule
A-GEM    & & lr : 0.05 (cifar), 0.1 (minImg)   \\
       & & memory size (samples) : 2000 (cifar), 500 (minImg)  \\
\midrule
ER\_{Res}    & & lr : 0.05 (cifar), 0.1 (minImg)   \\
       & & memory size (samples) : 2000 (cifar), 500 (minImg) \\
\midrule
EWC    & & lr : 0.03 (minImg, superclass), 0.05 (cifar)  \\
       & & regularization coefficient : 1000 (superclass), 5000 (cifar, minImg)  \\
\midrule
HAT    & & lr : 0.03 (minImg), 0.05 (cifar) \\
       & & $s_{max}$ : 400 (cifar, minImg)  \\
       & & $c$ : 0.75 (cifar, minImg)  \\
\midrule
Multitask & & lr : 0.05 (cifar), 0.1 (minImg)  \\
\midrule
GPM  & & lr : 0.01 (cifar, superclass), 0.1 (minImg)  \\
& & $n_s$ : 100 (minImg), 125 (cifar, superclass)  \\
\midrule
TRGP  & & lr : 0.01 (cifar, superclass), 0.1 (minImg)  \\
& & $n_s$ : 100 (minImg), 125 (cifar, superclass)  \\
& & $\epsilon^l$ : 0.5 (cifar, miniImg, superclass) \\
& & \# tasks for trust region : Top-2 (cifar, miniImg, superclass) \\
\midrule
FS-DGPM  & & lr, $\eta_3$ : 0.01 (cifar, superclass) \\
& & lr for sharpness, $\eta_1$ : 0.001 (cifar), 0.01 (superclass) \\
& & lr for DGPM, $\eta_2$ : 0.01 (cifar, superclass) \\
& & memory size (samples) : 1000 (cifar, superclass) \\
& & $n_s$ : 125 (cifar, superclass)  \\
\midrule
\textbf{SGP (ours)}  & & lr : 0.01 (superclass),  0.05 (cifar), 0.1 (minImg)  \\
& & $n_s$ : 100 (minImg), 125 (cifar, superclass)  \\
& & $\alpha$ : 1 (minImg), 3 (superclass), 5, 10 (cifar), 15, 25 \\
\bottomrule
\end{tabular}}
\caption{List of hyperparameters in our method and baseline approaches. Here, `lr' represents (initial) learning rate. In the table we represent Split CIFAR-100 as `cifar', Split miniImageNet as `minImg' and CIFAR-100 Superclass as 'superclass'.}
\label{tab:app_tab2}
\end{table*}
\vspace{-10pt}
\subsection{Architecture Details}
For \textbf{Split CIFAR-100} experiments, similar to Saha et al.~\citeyearpar{gpm}, we used a 5-layer AlexNet network with 3 convolutional layers having 64, 128, and 256 filters with $4\times 4$, $3\times 3$, and $2\times 2$ kernel sizes respectively, followed by two fully connected layers having 2048 units each. Each layer has batch normalization except the classifier layer. We use $2\times 2$ max-pooling after the convolutional layers. Dropout of 0.2 is used for the first two layers and 0.5 for the rest. For \textbf{Split miniImageNet} experiments, we use a reduced ResNet18 architecture (Saha et al.~\citeyear{gpm}) with 20 as the base number of filters. We use convolution with stride 2 in the first layer and $2\times 2$ average-pooling before the classifier layer. For \textbf{CIFAR-100 Superclass} experiments, similar to~\citet{apd}, we use a modified LeNet-5 architecture with 20-50-800-500 neurons. All the networks use ReLU in the hidden units and softmax with cross-entropy loss in the final layer. No bias units are used and batch normalization parameters are learned for the first task and shared with all the other tasks (Saha et al.~\citeyearpar{gpm}). Each task uses a separate classifier head (multi-head setup) which is trained without any gradient projection. 
\vspace{-20pt}
\subsection{Threshold Hyperparameter ($\epsilon_{th}$)}
We use the same threshold hyperparameters as GPM (Saha et al.~\citeyear{gpm}) in all the experiments. For split CIFAR-100 experiment, we use $\epsilon_{th}=0.97$ for all the layers and increase the value of $\epsilon_{th}$ by $0.003$ for each new task. For Split miniImageNet experiment, we use $\epsilon_{th}=0.985$ for all the layers and increase the value of $\epsilon_{th}$ by $0.0003$ for each new task. For CIFAR-100 Superclass experiment, we use $\epsilon_{th}=0.98$ for all the layers and increase the value of $\epsilon_{th}$ by $0.001$ for each new task.

\subsection{List of Hyperparameters}
A list of hyperparameters in our method and baseline approaches is given in Table~\ref{tab:app_tab2}.

\subsection{SGP Implementation: Software, Hardware and Code}
We implemented SGP in~\texttt{python(version 3.9.13)} with~\texttt{pytorch(version 1.11.0)} and~\texttt{torchvision(version 0.12.0)} libraries. For reinforcement learning tasks we additionally used~\texttt{gym(version 0.21.0)},~\texttt{gym-atari(version 0.19.0)} and~\texttt{stable-baselines3(version 1.6.0)} packages. We ran the codes on a single NVIDIA A40 GPU (CUDA~\texttt{version 11.7}) and reported the results in the paper. Codes will be available at \href{https://github.com/sahagobinda/SGP}{\textcolor{magenta}{\texttt{https://github.com/sahagobinda/SGP}}}.

\section{Additional Results and Analyses: Continual Image Classification Tasks}\label{App_add_results_1}

\begin{figure*}[t]
\begin{centering}
  \includegraphics[width=0.82\textwidth,keepaspectratio,page=7]{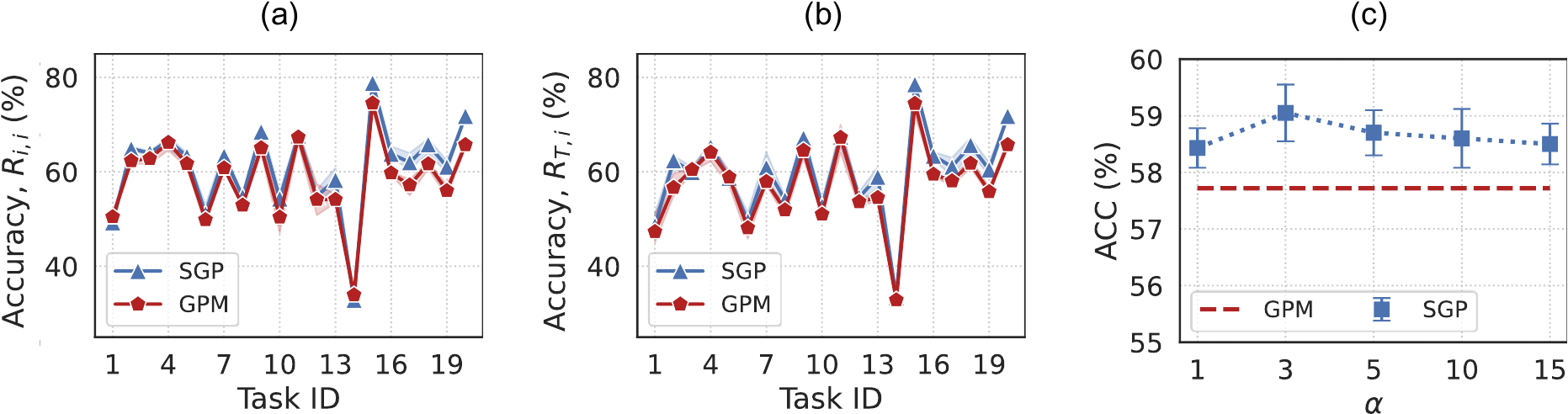}
  \caption{Test accuracy of each CIFAR-100 Superclass task (a) after learning that task, (b) after learning all the tasks. (c) Impact of varying $\alpha$ on SGP performance for CIFAR-100 Superclass tasks.} 
\label{fig:app_fig1}
\end{centering}
\vspace{-5pt}
\end{figure*}

\subsection{Task-Level Generalization and Forward Transfer}
In Figure~\ref{fig:app_fig1}(a) we plot $R_{i,i}$, which is the test accuracy of $i^{th}$ task immediately after learning $i^{th}$ task during CL process for CIFAR-100 Superclass tasks. Here we find that relative forward transfer between SGP and GPM, $\Delta_{SGP-GPM}$ is approximately $+2\%$. This results again show better task-level generalization is achieved with SGP. Next, in Figure~\ref{fig:app_fig1}(b) we plot $R_{T,i}$, which is final test accuracy of each task after learning all $T$ tasks for CIFAR-100 Superclass. Here we observe that in almost all the tasks SGP has higher final accuracy than GPM which translates to better average accuracy (ACC) in SGP. 

\subsection{Training Time Measurements and Comparisons}
Training times for all the tasks in sequence for different experiments are measured on a single NVIDIA A40 GPU. Training time comparisons among different methods are given in Table~\ref{tab:app_time} where time is normalized with respect to the value of GPM.    
\renewcommand{\arraystretch}{1.1}
\begin{table*}[t]\centering
\centering
\vspace{-10pt}
\scalebox{0.8}{
\begin{tabular}{@{}lccccccccc@{}}
&  \multicolumn{5}{c}{}  \\
\toprule
&  \multicolumn{9}{c}{\textbf{Methods}} \\ 
\cmidrule{2-10}
Dataset &  \multicolumn{1}{c}{OWM} & \multicolumn{1}{c}{EWC} & \multicolumn{1}{c}{HAT}& \multicolumn{1}{c}{A-GEM}&  \multicolumn{1}{c}{ER\_Res}  & \multicolumn{1}{c}{GPM} & \multicolumn{1}{c}{TRGP} & \multicolumn{1}{c}{FS-DGPM} & \multicolumn{1}{c}{\textbf{SGP (ours)}}\\
\midrule
Split CIFAR-100    &2.41 &1.76 &1.62 &3.48 &1.49 &1 &1.65& 10.25 & 1.16\\
Split miniImageNet  &- &1.22 &0.91 &1.79 &0.82 &1 &1.78& - &1.05\\
CIFAR-100 Superclass  &- &- &- &- &- &1 &1.52& 11.03 &1.17\\
\bottomrule
\end{tabular}}
\caption{\textcolor{black}{Training time comparison on different datasets. Here the training time is normalized with respect to the value of GPM~(Saha et al.~\citeyear{gpm}; Lin et al.~\citeyear{trgp}).}}
\label{tab:app_time}
\vspace{-10pt}
\end{table*}

\subsection{Impact of Varying $\alpha$}
Impact of varying $\alpha$ on SGP performance (ACC) for CIFAR-100 Superclass dataset is shown in Figure~\ref{fig:app_fig1}(c). Best performance obtained at $\alpha=3$ and SGP performance approaches GPM result for larger $\alpha$ values. 

\subsection{Basis Importance Distributions}
\begin{figure*}[t]
\begin{centering}
  \includegraphics[width=0.95\textwidth,keepaspectratio,page=8]{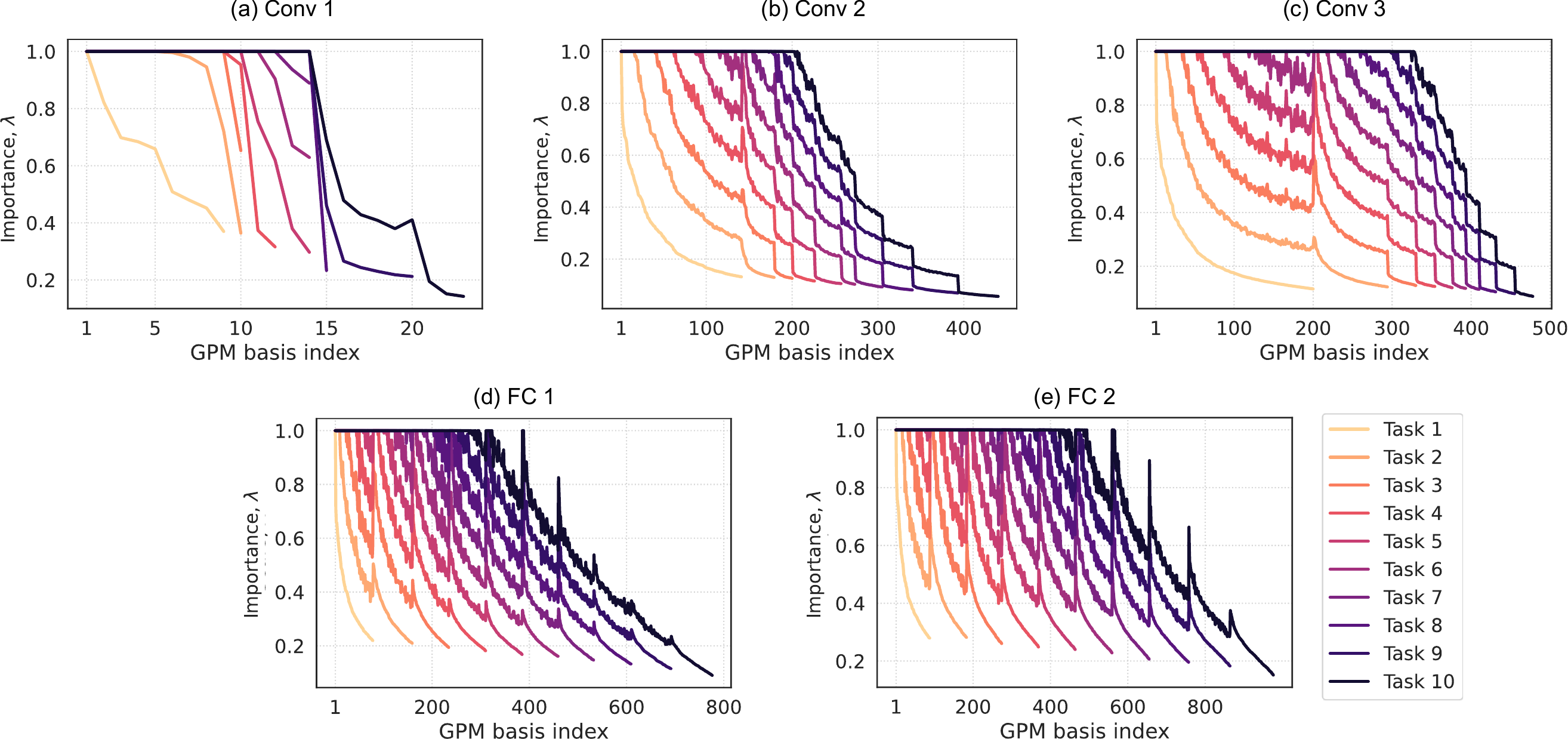}
  \caption{Basis importance ($\bm{\lambda}$) distributions at different layers of the (AlexNet) network for different tasks from Split CIFAR-100 dataset. } 
\label{fig:app_imp1}
\end{centering}
\vspace{-5pt}
\end{figure*}

\begin{figure*}[t]
\begin{centering}
  \includegraphics[width=0.7\textwidth,keepaspectratio,page=9]{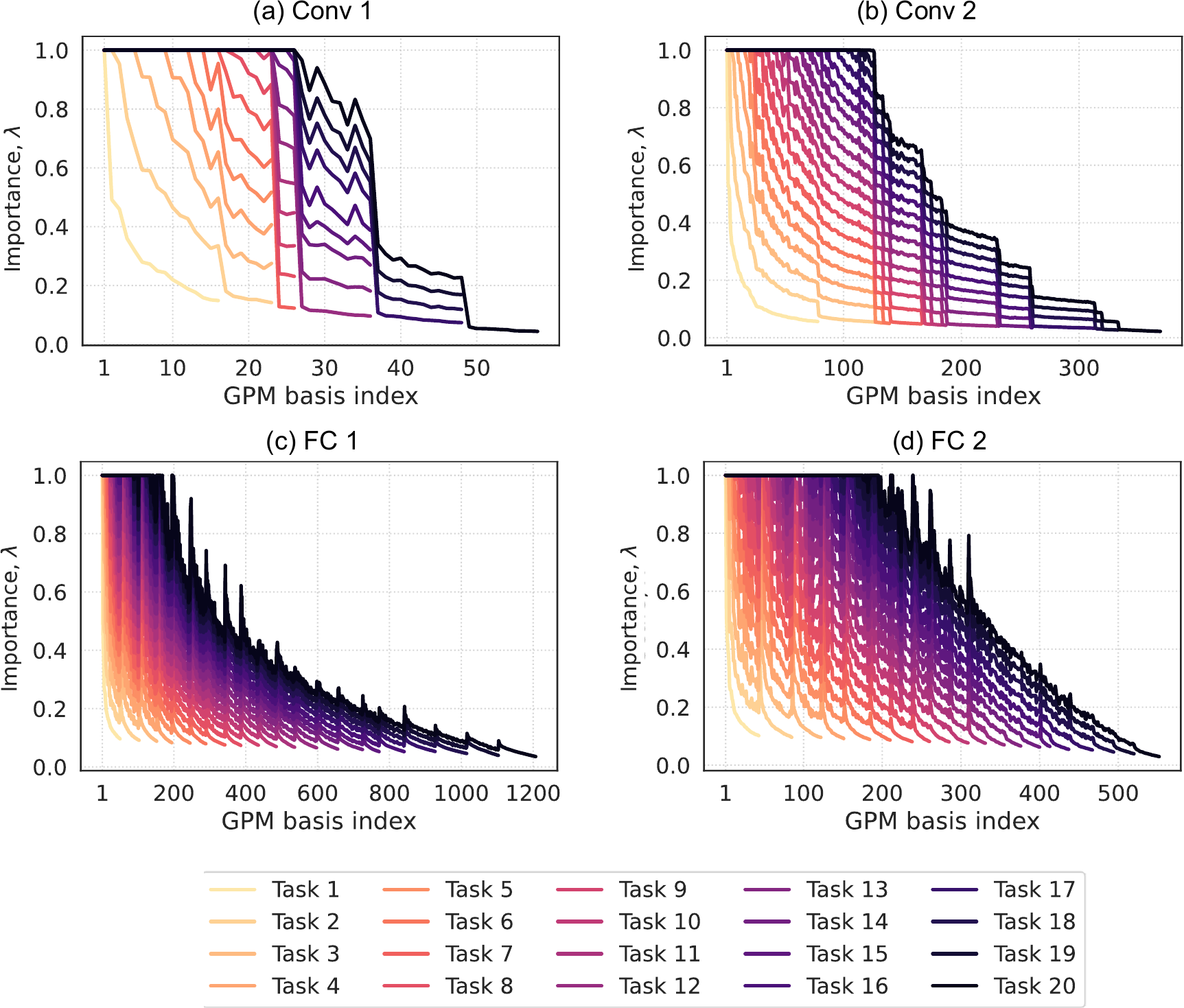}
  \caption{Basis importance ($\bm{\lambda}$) distributions at different layers of the (LeNet) network for different tasks from CIFAR-100 Superclass dataset.} 
\label{fig:app_imp2}
\end{centering}
\vspace{-5pt}
\end{figure*}

Basis importance ($\bm{\lambda}$) distributions at different layers of the networks for different tasks from Split CIFAR-100 dataset and CIFAR-100 Superclass dataset are given in Figure~\ref{fig:app_imp1} and Figure~\ref{fig:app_imp2} respectively.

\clearpage

\begin{figure*}[t]
\begin{centering}
  \includegraphics[width=0.95\textwidth,keepaspectratio,page=10]{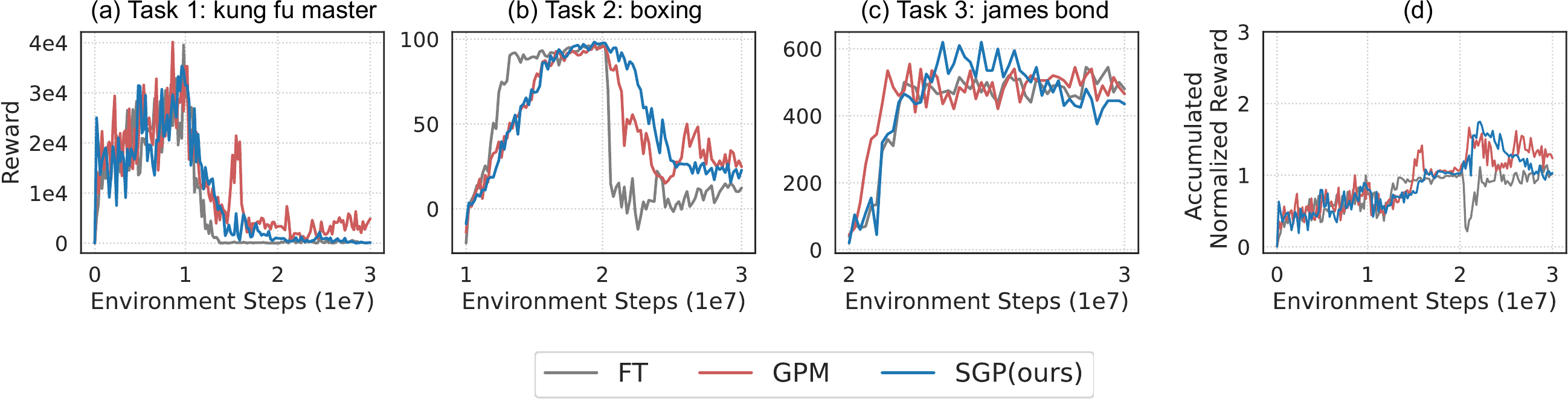}
  \vspace{-2mm}
  \caption{An example continual reinforcement learning scenario with 3 tasks (games) illustrates the catastrophic forgetting in GPM and SGP when projected gradients are passed to the \texttt{Adam} optimizer for model update. (a)-(c) Rewards on 3 sequential Atari games (tasks). Each task is trained for 10 million environment steps, and 30 million environment steps are used in total. Both GPM and SGP show catastrophic forgetting. Thus, (d) accumulated normalized rewards in GPM and SGP oscillate around 1 similar to FT baseline.} 
\label{fig:app_adam_fail}
\end{centering}
\vspace{-5pt}
\end{figure*}

\section{Additional Results and Analyses: Continual Reinforcement Learning Tasks}\label{App_add_results_2}
\subsection{Modified Adam Algorithm with Gradient Projection}
The previous gradient projection based continual learning algorithms (Zeng et al.~\citeyear{owm}; Farajtabar et al.~\citeyear{ogd}; Saha et al.~\citeyear{gpm}; Deng et al.~\citeyear{fs-dgpm}; Lin et al.~\citeyear{trgp}) used plain SGD as the network optimizer. None of them reported results with other optimizers such as~\texttt{Adam}~\cite{adam} that could be needed for application domains other than image classification. For instance, in reinforcement learning tasks when we pass the projected gradients (Equation~\ref{eq:sgp}) from GPM/SGP to~\texttt{Adam} it modifies the gradients in a way (Line 9-12 in Algorithm~\ref{alg:sgp_adam}) that fails to preserve the gradient orthogonality/scaled projection conditions required for CL. As a result, such gradient updates show catastrophic forgetting. An example scenario is provided in Figure~\ref{fig:app_adam_fail}, where 3 tasks (games) are learned sequentially with ~\texttt{Adam} optimizer by GPM, SGP and FT agents. All these methods forget old games as they learn a new one, thus their accumulated normalized reward oscillates around 1 (Figure~\ref{fig:app_adam_fail}(d)). To resolve this issue, we modify~\texttt{Adam} in Algorithm~\ref{alg:sgp_adam} which we denote as~\texttt{Adam-GP}. Here, output of ~\texttt{Adam} optimizer (Line 12) is processed with gradient projection algorithms, and model is updated with the projected gradients (Line 13-16). With such modification, we obtain SOTA performance in continual reinforcement learning tasks (Section~\ref{clrl}). Thus, we believe~\texttt{Adam-GP} will be beneficial for expanding the application domains of the gradient projection based CL algorithms.      

\begin{algorithm}
   \caption{Modified Adam Algorithm for Continual Learning with Gradient Projection}
   \label{alg:sgp_adam}
    \begin{algorithmic}[1]
       \Procedure{Adam-GP}{$f(\theta)$, $\beta_1$, $\beta_2$, $\epsilon$, $\eta$, $\mathcal{M}$, $\mathcal{S}$}
       \Statex \textbf{Inputs}: $f(\theta)$: Stochastic objective function with parameters $\theta$ (a neural network in our case) , $\beta_1,\beta_2\in[0,1)$: Exponential decay rates for the moment estimates, $\eta$: Stepsize, $\epsilon=10^{-8}$, $\theta_0$: Initial parameter vector, $\mathcal{M}$ : Basis memory (GPM), $\mathcal{S}$: Basis importance  
      \State $m_0 \gets 0$  \Comment{Initialize $1^{st}$ moment vector}
      \State $v_0 \gets 0$  \Comment{Initialize $2^{nd}$ moment vector}
      \State $t \gets 0$  \Comment{Initialize timestep}
      
      \While{$\theta_t$ not converged}
        \State $t \gets t+1$
        \State $g_t \gets \nabla_\theta f_t(\theta_{t-1})$ \Comment{gradients w.r.t. stochastic objective at timestep $t$ (for the current task)}
        \State $m_t \gets \beta_1 . m_{t-1} + (1-\beta_1) . g_t$ \Comment{Update biased first moment estimate}
        \State $v_t \gets \beta_2 . v_{t-1} + (1-\beta_2) . g_t^2$ \Comment{Update biased second raw moment estimate, $g_t^2$ denotes element-wise square }
        \State $\hat{m_t} \gets \frac{m_t}{(1-\beta_1^t)}$ \Comment{Compute bias-corrected first moment estimate, $\beta_1^t$ denotes $\beta_1$ to the power $t$ }
        \State $\hat{v_t} \gets \frac{v_t}{(1-\beta_2^t)}$ \Comment{Compute bias-corrected second raw moment estimate, $\beta_2^t$ denotes $\beta_2$ to the power $t$ }
        \State $g_t^{Adam} \gets \frac{\hat{m_t}}{\sqrt{\hat{v_t}}+\epsilon}$ \Comment{\textbf{gradient output from \texttt{Adam}}}
        \State $\bm{M} \gets \texttt{getmatrix}(\mathcal{M})$ \Comment{get the basis matrix}
        \State $\bm{\Lambda} \gets \texttt{getmatrix}(\mathcal{S})$ \Comment{get the basis importance matrix, for GPM $\bm{\Lambda}=\bm{I}$ is Identity matrix}
        \State $g_t^{GP} \gets g_t^{Adam} - (\bm{M\Lambda M^T})g_t^{Adam}$ \Comment{get the \textbf{final projected gradient}, layerwise operation see Section~\ref{sec:sgp_method}}
        \State $\theta_t \gets \theta_{t-1} - \eta g_t^{GP} $ \Comment{update the model with projected gradient}
        
    \EndWhile
    \State \textbf{return} $f(\theta)$
    \EndProcedure
    \end{algorithmic}
\end{algorithm}

\end{document}